\documentclass[10pt,conference]{ieeeconf}
\usepackage[utf8]{inputenc}
\usepackage{amsmath}
\usepackage{amssymb}
\usepackage{graphicx}
\usepackage{fancyvrb}

\begin{document}
\title{A Constructive Approach for One-Shot Training of Neural Networks Using Hypercube-Based Topological Coverings }
\author{W. Brent Daniel, Enoch Yeung}
\maketitle

\thispagestyle{plain}
\pagestyle{plain}





\begin{abstract}
In this paper we presented a novel constructive approach for training deep neural networks using geometric approaches.   We show that a topological covering can be used to define a class of distributed linear matrix inequalities, which in turn directly specify the shape and depth of a neural network architecture.  The key insight is a fundamental relationship between linear matrix inequalities and their ability to bound the shape of data, and the rectified linear unit (ReLU) activation function employed in modern neural networks.    We show that unit cover geometry and cover porosity are two design variables in cover-constructive learning that play a critical role in defining the complexity of the model and generalizability of the resulting neural network classifier.  In the context of cover-constructive learning, these findings underscore the age old trade-off between model complexity and overfitting (as quantified by the number of elements in the data cover) and generalizability on test data.  Finally, we benchmark on algorithm on the Iris, MNIST, and Wine dataset and show  that the  constructive  algorithm  is  able  to  train  a  deep  neural network  classifier  in  one  shot, achieving  equal or superior levels of training and test classification accuracy with  reduced training  time.  
\end{abstract}

\section{Introduction}

Artificial neural networks have proven themselves to be useful, highly flexible tools for addressing many complex problems where first-principles solutions are infeasible, impractical, or undesirable. They have been used to address challenging classification problems ranging from wine typing to complex image analysis, voice recognition, language translation, and beyond.

The same flexibility that allows neural networks to be applied in such disparate contexts, however, can also lead to ambiguity in their appropriate definition and training. Deep neural networks, for example, are composed of multiple hidden layers with each hidden layer containing many nodes, each completely connected to the nodes in the preceding layer by a set of weights $\mathbf{W}_h$. Historically there has not been a functional relationship or algorithmic approach that allows researchers to define or derive a neural network's structural characteristics from either the problem specification or the associated training data. A neural network's topology \emph{can} be optimized for a given problem, but this effectively results in a nested series of optimizations with the outer optimization steps tasked with incrementally assessing the most effective network topology \cite{Lam2001}.

Similarly, there has been no \emph{a priori} way to specify reasonable initial values for the weights, $\mathbf{W}_h$. In practice these weights are stochastically initialized. The training process then optimizes their values so as to maximize network performance against a specified metric or cost function. This cost function can have complex dependencies on the input parameters. Many algorithms, especially those that rely on gradient information, can become stuck in local minima, limiting the predictive quality of a network for a given training instance. The result is that the same structural topology and training data can yield neural networks with a broad distribution of predictive qualities from one training run to the next. These stochastic effects can be most marked when the volume of training data is relatively small, yielding an optimization problem with relatively few constraints compared to the dimensionality of the parameter space. Such effects can be reduced by the choice of training algorithm, its parameterization, or by repeated training restarts, but this correspondingly increases the computational complexity and training time. Additionally, it's typically impossible to unambiguously specify a finite stop condition for training. This is the result of three factors: the training process is stochastic, the metric space has the potential for local minima, and the global minimum value is unknown beforehand.

In this paper we introduce a constructive method for the design of neural networks that rely on geometric representation of the data. The algorithm directly addresses the issues outlined above, including, 1) providing a concise structural definition of the neural network and its topology, 2) assigning network connection weights deterministically, 3) incorporating approximations that allow the algorithm to construct neural networks that in many cases have greater mean accuracy and better precision than traditionally trained networks, especially when training data is relatively sparse,  
4) having a well-defined stop condition for training, and
5) inherently providing a clear interpretation of what and how information is encoded within the resulting neural network.

\section{Constructive Learning Using Topological Covers}
In what follows, we introduce a three-step approach for constructive training of a ReLU neural network:
\begin{itemize}
\item One or more topological covering maps are defined between a network's desired input and output spaces;
\item These covering spaces are encoded as a series of linear matrix inequalities; and, finally,
\item The series of linear matrix inequalities is translated into a neural network topology with corresponding connection weights.
\end{itemize}
Many schemes for defining topological covering maps can be imagined, each with their own strengths and weaknesses. In the present work a relatively simple approach has been devised that prizes computational efficiency and classification efficacy under sparse training data. The resulting algorithm scales roughly linearly in the number of training data points and only weakly with the number of input dimensions. This need not be the case. We'll first discuss a similar, but brute force approach, where memory requirements and computational complexity scale exponentially with input dimension.

\subsection{Covering with Uniform Hypercubes}

Let's let $\mathbf{X} \in \mathbb{R}^n$ and $\mathbf{Y} \in \mathbb{R}^m$ denote the potential input and output spaces of a neural network, where $n$ is the input dimension and $m$ is the number of classification categories. The neural network then performs a mapping/transformation, $f: \mathbf{X} \rightarrow \mathbf{Y}$, in such a way that an output vector, $\mathbf{y} \in \mathbf{Y}$, is, hopefully, more clearly and trivially indicative of membership within the appropriate category than the original point, $\mathbf{x} \in \mathbf{X}$. Let's assume that each potential training vector, $\mathbf{x}_i$, has an associated label, $L_i$, that indicates its membership in a category, $c \in C$, where $C = \{c_1, \ldots, c_m \}$. The set of points belonging to category, $c$, is then given by $\mathbf{X}^c = \{ \mathbf{x}_i \in \mathbf{X} \; | \; L_i = c \}$.

The set of training data corresponding to a single category of input, $\mathbf{X}^c$, can be thought of as a point cloud in an $n$-dimensional space.  This point cloud can be characterized by its maximal and minimal extent, $b_j^<$ and $b_j^>$, in each of the $n$ dimensions, where $j \in \{1 \ldots n \}$. This set of $2n$ constraints is sufficient to define a bounding hypercube represented by a collection of $(n-1)$-dimensional hyperplanes, each plane being orthogonal to one of the $n$ directions. Defining $m$ such bounding hypercubes would allow one to rapidly characterize regions of the input parameter space corresponding to each category.

For complex classification problems, however, it's likely that portions of these bounding hypercubes may overlap, leading to apparent ambiguities in the classification of both training and evaluation data. This scheme could, however, be refined to reduce the potential for ambiguous classifications. Instead of defining one hypercube per classification category, let's define a more general set of hypercubes that will be assigned a classification category based on \emph{a posteriori} point membership. Let's let $N_T$ denote the total number of training points across all categories. The full set of points, $\mathbf{X}^T = \{ \mathbf{x}_i \; | \; i \in \{1 \ldots N_T \}\}$, has bounds $b_j^{T<}$ and $b_j^{T>}$ for $j \in \{1 \ldots n \}$. These are is sufficient to define a bounding hypercube, $H_T$.

Let's assume that the smallest point cloud feature that should be resolved is of length scale $l$, where $l$ denotes a distance in the parameter space. The full hypercube, $H_T$, may then be subdivided into a set of smaller hypercubes, $\mathbf{H}$, by subdividing each axis of $H_T$ into bins of width, $l$, with each of the resulting hypercubes having volume $l^n$. The set of cubes, $\mathbf{H}$, represents a complete covering of $H_T$ with no overlapping regions and no empty spaces. The set of points that lies within one of these hypercubes, $H_k$, is given by:
\begin{equation}
\mathbf{X}_k = \{ \mathbf{x}_i \; | \; \forall j \; (b_{kj}^< \le x_{ij} < b_{kj}^>) \}
\end{equation}
where, for example, $b_{kj}^<$ denotes the lower boundary in the $j$-th direction of the $k$-th hypercube. The corresponding set of categories represented by the points in $H_k$ is:
\begin{equation}
C_k = \{ L_i \; | \; \forall j \; (b_{kj}^< \le x_{ij} < b_{kj}^>) \}
\end{equation}
A category may then be assigned to each $H_k$ based on the class membership, $C_k$, of the points that lie within its bounds. If the hypercube contains no points, $C_k = \emptyset$, it is marked as empty; if all points are of the same class, $(\exists c)(\forall i) (C_{ki} = c )$, the hypercube is assigned the category, $c$; otherwise, if the hypercube contains points of multiple categories, its classification remains ambiguous.

If the length scale, $l$, is chosen to be sufficiently small, each member, $H_k$, of the set of hypercubes can be classified unambiguously as either empty or containing points from a single class. Let's let $Z^c$ indicate the indices of points in category $c$, $Z^c = \{i | L_i = c \}$. The minimum distance between two points of different categories is, then:
\begin{equation}
l^{*} = \min_{a \in C, b \in C, a \ne b} \left( \min_{i \in Z^a, j \in Z^b} |\mathbf{x}_i - \mathbf{x}_j| \right)
\end{equation}
So long as $l$ is chosen roughly such that $l < l^{*}$, an unambiguous covering can be defined.

Operationally, the challenge with this approach is that the number of hypercubes, $K$, required to uniformly cover $H_T$ scales poorly with both minimum feature size, $l$, and the dimension of the parameter space, $n$; that is: $K \propto (1/l)^n$, where $(1/l)$ is proportional to the number of divisions along each axis of $H_T$. For example, even if $(\forall j) \; [(b_j^{T>} - b_j^{T<})/l \sim 10]$ and $n=32$, both relatively small numbers for real-world problems, this brute force approach is essentially computationally intractable, requiring $O(10^{32})$ hypercubes.

\subsection{Hypercube Bisection}

The inefficiency of the above approach arises from the need to cover $H_T$ completely with hypercubes of uniform dimension. Even if there are features as small as length scale, $l^*$, it doesn't necessarily follow that that level of fidelity is required everywhere. In fact, one could argue that such a level of fidelity is only critical where it is required to delineate the boundary between different classification regions.

An alternative approach would, thus, be to adaptively subsect the hypercube, $H_T$, in such a way as to create a covering of the point cloud regions with a minimal---or at least a relatively small---number of subsections. A practicable, reasonably efficient algorithm for accomplishing this is presented here. The bounding hypercube, $H_T$, containing the full set of training points, $\mathbf{X}^T$, is first determined. This hypercube is then recursively bisected following a simple set of rules until no more inhomogeneous hypercubes that are unconstrained remain. As the algorithm proceeds, four different sets of hypercubes are tracked: those that contain a homogeneous distribution of point types, $\mathbf{H}^H$; those that contain an inhomogeneous distribution of point types whose further bisection would not violate any constraints, $\mathbf{H}^I$; those inhomogeneous cubes whose further bisection would violate a constraint, $\mathbf{H}^V$; and those that are empty, $\mathbf{H}^E$.

The algorithm is initiated with $\mathbf{H}^I = \{ H_T \}$ and $\mathbf{H}^H = \mathbf{H}^V = \mathbf{H}^E = \emptyset$. The set of inhomogeneous cubes, $\mathbf{H}^I$, represents the algorithm's working stack. During each iteration, while $\mathbf{H}^I \ne \emptyset$, a hypercube with an inhomogeneous distribution of point categories is pulled from $\mathbf{H}^I$. The axis along which the bisection of this hypercube would provide the most benefit, $\alpha^*$, is determined subject to a constraint on the minimum length scale of the daughter hypercubes. Bisection along an axis, $\alpha$, is forbidden if the corresponding dimension of the daughter hypercubes, $H_{(a), \alpha}$ and $H_{(b), \alpha}$, would be of length less than, $l$.

The `best' bisection axis is determined by the degree to which bisection along that axis would homogenize the distribution of the point classes in the resulting daughter hypercubes. To determine the best bisection axis a test bisection is performed along each axis, $\alpha$, and the points of the parent hypercube assigned to each of the daughters based on their spatial location. The degree of homogeneity of the points within each daughter is then computed. Let's denote the indices of points in the class, $c$, within a daughter hypercube, $H_{(a), \alpha}$, by $Z_{(a)}^c$, and the number of such points by $|Z_{(a)}^c|$. A minimum homogeneity among point class pairs can be defined as:
\begin{equation}
h_{(a), \alpha}^< = \min_{c \in C, c^\prime \in C, c \ne c^\prime} f \left( Z_{(a)}^c, Z_{(a)}^{c^\prime} \right)
\end{equation}
where
\begin{equation}
f \left( Z_{(a)}^c, Z_{(a)}^{c^\prime} \right) = 0,
\end{equation}
if $|Z_{(a)}^c| = |Z_{(a)}^{c^\prime}| = 0$; and 
\begin{equation}
f \left( Z_{(a)}^c, Z_{(a)}^{c^\prime} \right) = (|Z_{(a)}^c| - |Z_{(a)}^{c^\prime}|) / (|Z_{(a)}^c| + |Z_{(a)}^{c^\prime}|),
\end{equation}
otherwise. The greater minimum homogeneity of the two daughter hypercubes is, then, computed for each bisection axis, resulting in the following set of tuples:
\begin{equation}
\mathcal{H} = \{ (\alpha, \max(h_{(a), \alpha}^<, h_{(b), \alpha}^<)) \; | \; \alpha \in \{1 \ldots n \} \}
\end{equation}
A function, $g_i(.)$, is then defined that extracts the $i$-th element of a tuple. The maximum benefit of bisection is, $h^* = \max \, \{ g_2(h) \; | \; h \in \mathcal{H} \}$, and the axes of all bisections that would yield the maximum benefit,
\begin{equation}
\{ g_1(h) \; | \; h \in \mathcal{H} \wedge g_2(h) = h^* \}
\end{equation}
If the cardinality of this resulting set is one, that axis is selected for bisection. If the cardinality is greater than one, an axis of bisection, $\alpha^*$, is chosen at random from the set.

Note that by setting $f \left( Z_{(a)}^c, Z_{(a)}^{c^\prime} \right) = 0$ when $|Z_{(a)}^c| = |Z_{(a)}^{c^\prime}| = 0$, we have chosen to deemphasize the segregation of empty regions as a determining factor in choosing bisection axes. This need not be the case, but will largely be rendered a moot point when hypercube aspect ratio constraints are added in a following section.

If bisection along all potential axes would run afoul of the length scale constraint, the hypercube is placed into the $\mathbf{H}^V$ set and the algorithm repeated on the next member of $\mathbf{H}^I$. Otherwise, the parent hypercube is bisected into two daughter hypercubes along axis, $\alpha^*$. Each of the daughter hypercubes is, then, assigned to one of the sets, $\mathbf{H}^H$, $\mathbf{H}^I$, or $\mathbf{H}^E$, depending on the classes of the points that lie within it.

\begin{figure}
\center
\includegraphics[width=3.125in]{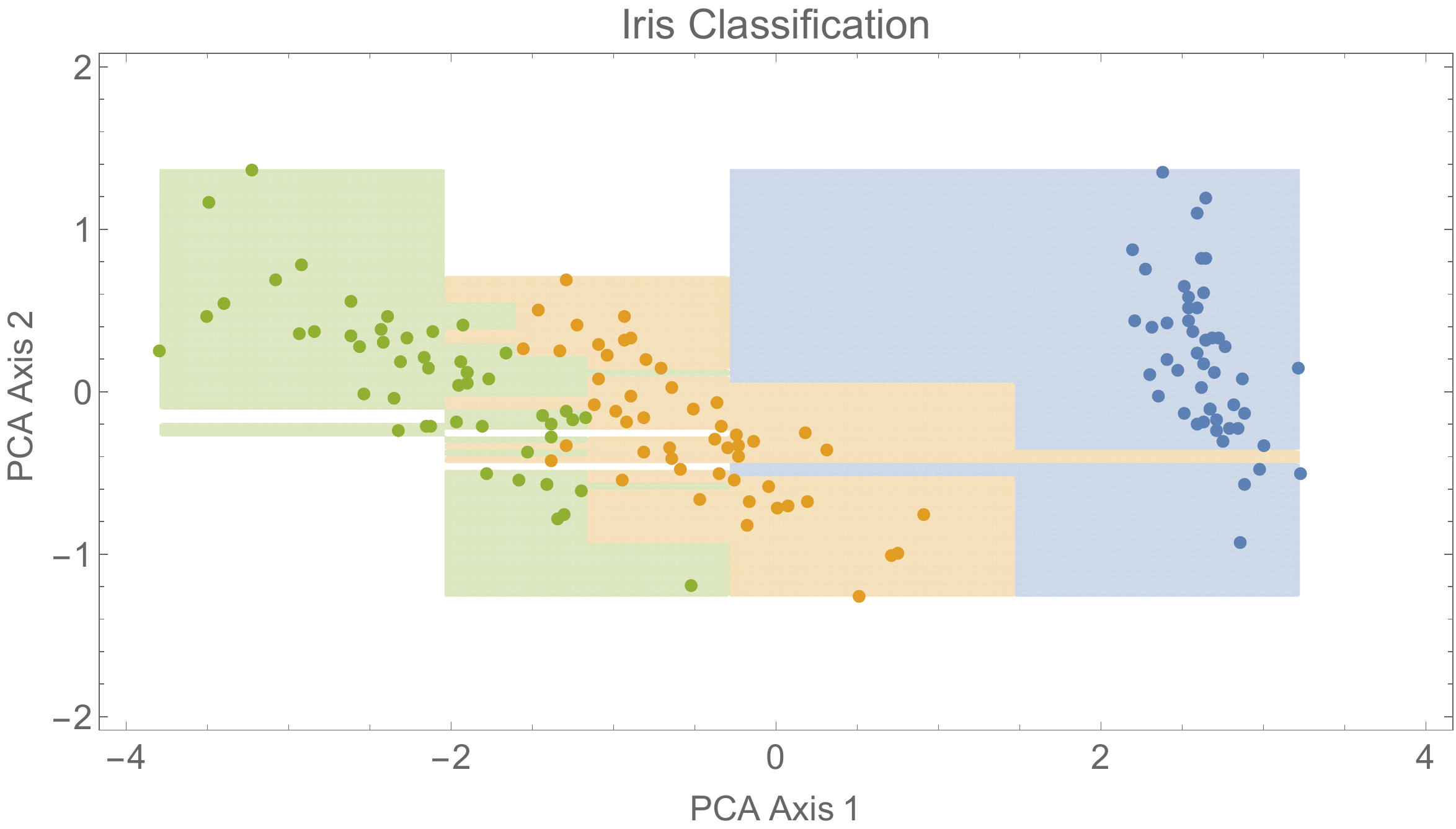}
\caption{The base bisection algorithm applied to point clouds of the first two principal component analysis (PCA) directions for parameters related to the classification of three categories of iris: setosa, versicolor, and virginia. Note that algorithm is unconstrained by hypercube aspect ratio, here, allowing for the development of narrow pancake-shaped hypercubes that may not generalize well to the classification of previously unseen data points.}
\label{fig:iris-base}
\end{figure}

The algorithm's use is illustrated in Fig.~\ref{fig:iris-base}, where it has been applied to a standard benchmark problem classifying iris species based on their morphological features \cite{Dua2017}. A principle components analysis (PCA) was done on the original four series included in the data set and only the two most important components retained for analysis. Two useful observations are readily made: first, the algorithm achieves 100\% accuracy in classifying the training data; second, one might suspect that the resulting covering would generalize poorly when used to classify previously unseen evaluation data. The latter is the result of two characteristics of the algorithm. First, because of the algorithm's sole focus on homogenization of the data points rather than regularization of the covering shapes, narrow peninsulas of one classification category's covering can often penetrate into other regions of the parameter space. Second, there are also large, sometimes inter-penetrating, regions of the relevant parameter space that have not been assigned a classification category at all. Both of these potential shortcomings will be addressed in a following section.

\subsection{Direct Restrictions on Cover Geometry to Avoid Overfitting}

As noted, the algorithm outlined above can lead to hypercubes with large aspect ratios, essentially high-dimensional pancakes whose extent along some direction(s) is vastly greater than along others. This is typically reflective of some underlying feature of the data that emphasizes bisection along a particular discriminatory axis. While this is of little consequence from the point of view of characterizing the training data, it can lead to poor performance when the resulting network is, then, asked to extrapolate to unseen data points.

One mechanism to address this issue is to limit the maximum aspect ratio that can be attained by a hypercube. An \emph{a posteriori} approach would be to have any bisection that results in the generation of daughter cubes with aspect ratios exceeding some threshold recursively trigger the bisection of those daughter hypercubes along orthogonal directions until the aspect ratio of every resulting daughter is within bounds. Unfortunately, this leads to nearly the same explosion in hypercube number that results from the initial uniform covering approach. Alternatively, the aspect ratio may be constrained \emph{a priori} so that bisection can occur only along directions such that the resulting daughters do not exceed a maximum aspect ratio, $r^*$. This alternative approach scales far better. It's slightly less efficient at developing coverings for training data than unconstrained bisection, in the sense that more optimal bisections are sometimes prevented. It results, however, in coverings with far fewer  artifacts that could subsequently lead to odd generalizations, and does so with minimal additional computational burden.

\begin{figure}
\center
\includegraphics[width=3.125in]{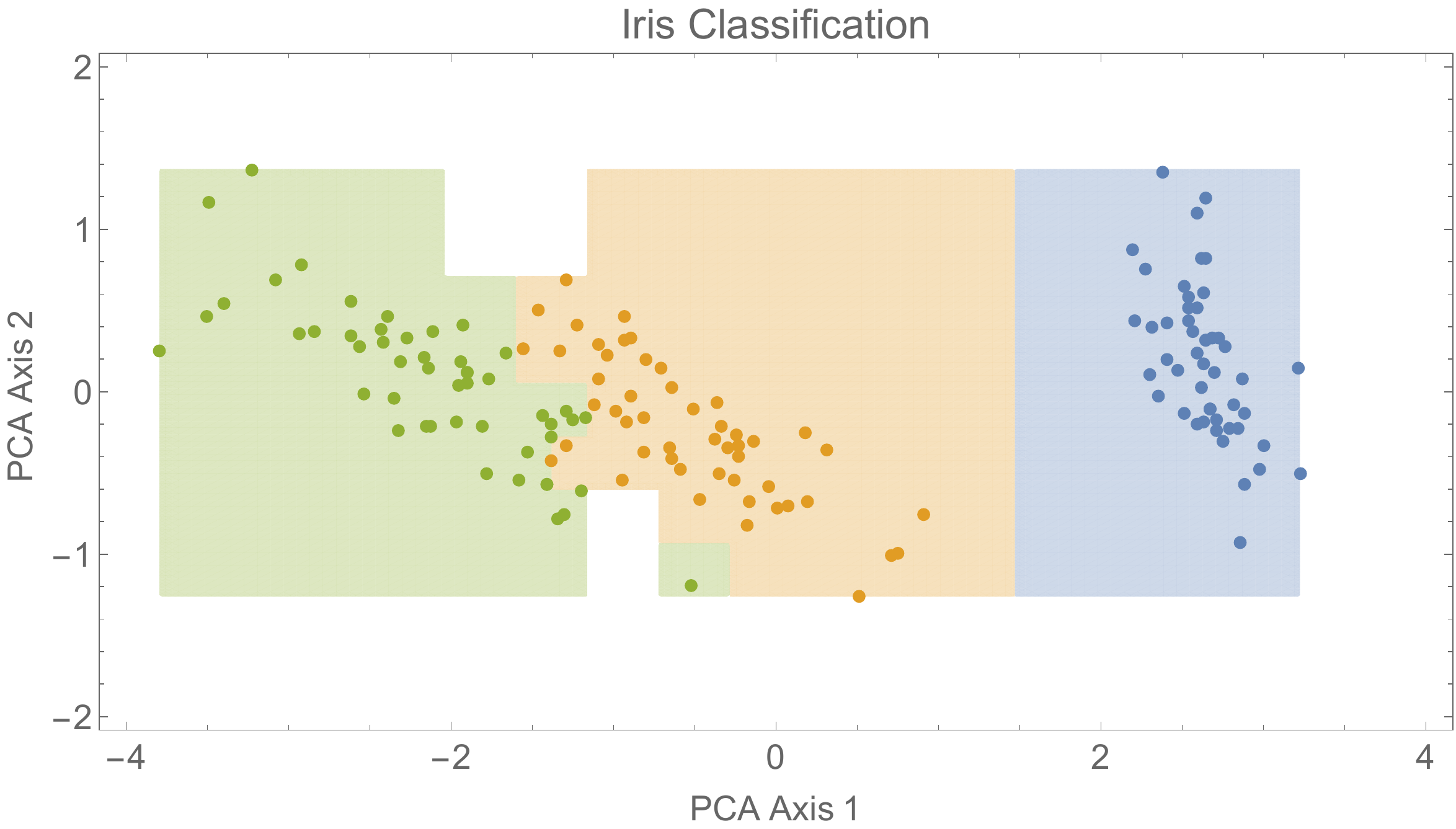}
\caption{Constraining daughter hypercube aspect ratio during each choice of bisection direction yields coverings that are more likely to generalize well, but which may still have holes in the covering.}
\label{fig:iris-aspect-ratio}
\end{figure}

If the breadths of a hypercube along each of its axes, $j$, are enumerated in a set, $B = \{ (b_j^> - b_j^<) \; | \; j \in \{1 \ldots n \} \}$, the hypercube's corresponding maximum aspect ratio is $r = \max B / \min B$. The algorithm above is, thus, updated such that viable bisection axes are limited to those along which bisection would not violate the constraint, $r \le r^*$, for the daughter hypercubes. The impact of incorporating such a constraint is illustrated in Fig.~\ref{fig:iris-aspect-ratio}. It should be noted that, in general---as well as specifically in the context of aspect ratio constraints---data sets should likely be normalized along each axis before analysis such that size and aspect ratio constraints are applied uniformly in all directions.

\subsection{Reducing Model Complexity via Classification Porosity}

As part of the bisection algorithm's implementation, the set of daughter hypercubes which are empty, $\mathbf{H}^E$, is tracked, and for efficiency reasons, removed from consideration for further bisection. As these cells contain no members of the original point cloud from which to assign a type, they essentially represent holes in the classification space. These holes partly reflect the underlying geometry of the original point cloud, but also contain relics of the bisection algorithm, including those resulting from random choices of bisection direction in the daughter hypercube's parental lineage. Both reduce the resulting neural network's ability to generalize classification knowledge beyond regions contained within the set of non-empty hypercubes in the topological covering.

\begin{figure}
\center
\includegraphics[width=3.125in]{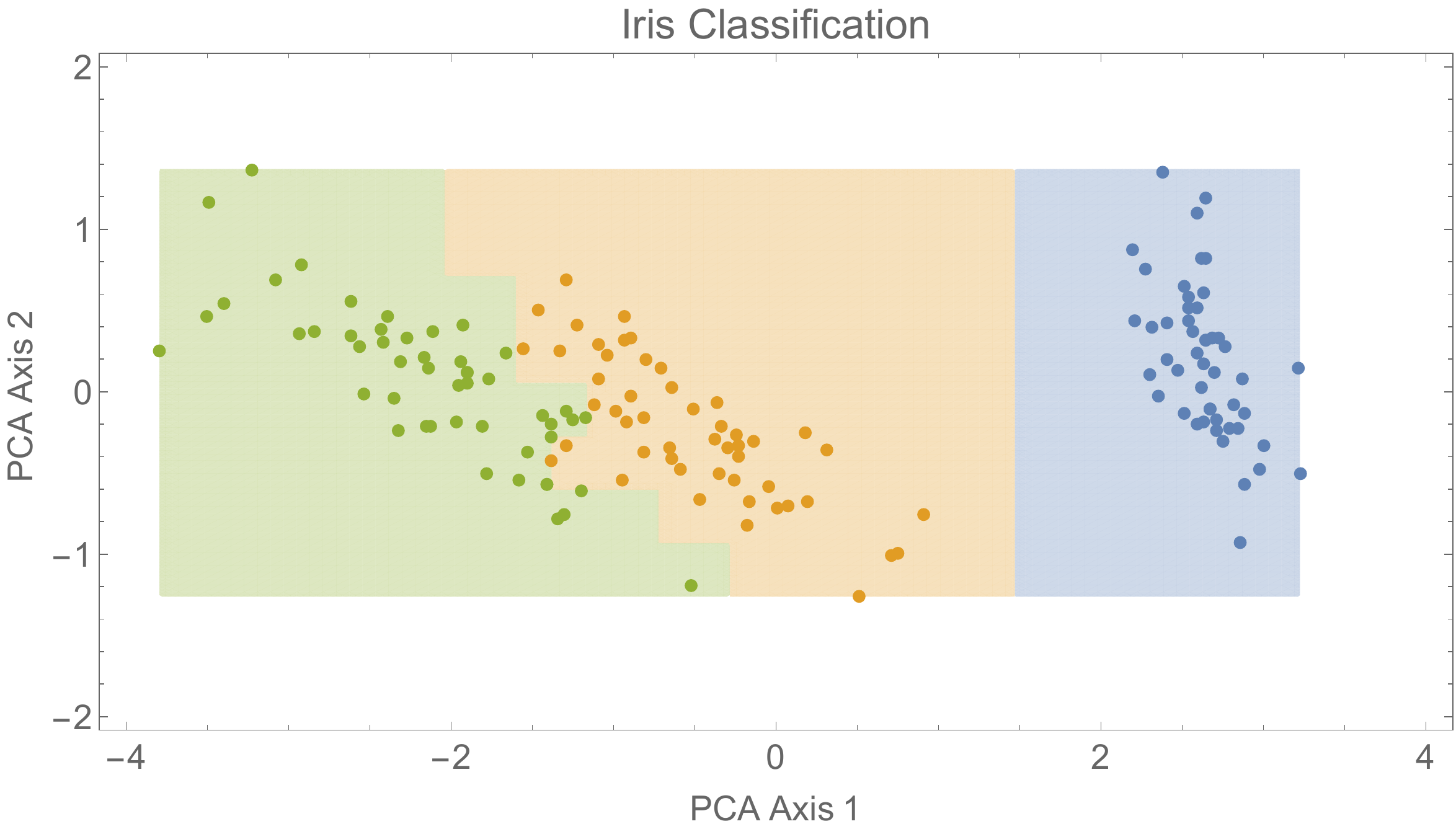} \\
\includegraphics[width=3.125in]{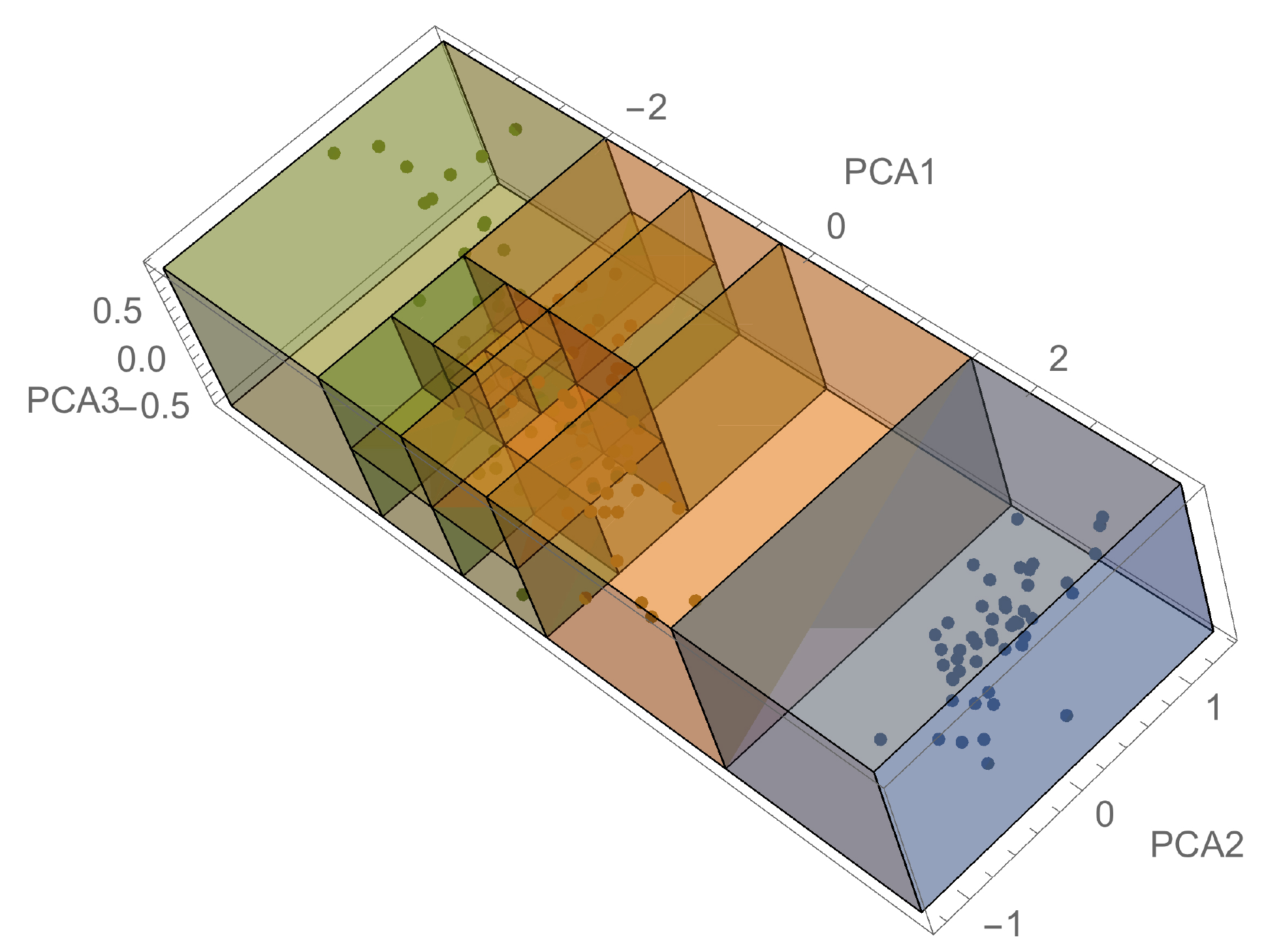}
\caption{Porous, empty regions of the classification/parameter space can be filled in using a recursive cellular automata based approach. The entire bisection algorithm scales well as the dimension, $d$, is increased. A 3D covering for iris classification based on the first three PCA directions is illustrated in the lower panel. The algorithm has been tested up to 32 dimensions. Computation time scales with the number of points in the point cloud rather than with the dimension.}
\label{fig:iris-porosity}
\end{figure}

How these empty regions are handled has a direct impact on network generalization. There is unlikely to be a single best approach, due to the no free lunch theorem. Currently, empty cells are recursively filled in using, what is effectively, a cellular automata schema. Empty cells bordering non-empty cells are assigned a classification category based on their total area of contact with non-empty cells of each type. The classification with maximum boundary area is assigned to the empty cell. As some empty cells may initially touch no non-empty cells, this process is repeated until all empty cells within $H^T$ have been assigned a classification category, yielding a non-porous covering. The covering that results after a category is assigned to empty hypercubes is shown in Fig.~\ref{fig:iris-porosity} for both two- and three-dimensional versions of the iris classification problem.

\subsection{Hypercubes As Linear Inequalities}

An advantage to using hypercubes as the building blocks of a topological covering is that they are convex $n$-gons with simple bounding hyperplanes. This convexity allows their interior to be delineated by a logical combination of inequalities, one per bound. Because the bounds are simple hyperplanes, the inequalities are all linear. For a given hypercube, $H_k$, the set of boundaries, $b_{kj}^<$ and $b_{kj}^>$, can, thus, be translated into a matrix of weights and a corresponding vector of biases:
\begin{align}
\mathbf{W}_k \mathbf{x} + \mathbf{v}_k \le 0,
\end{align}
where $\mathbf{W}_k \in \mathbb{R}^{2n \times n}$ and $\mathbf{v}_k \in \mathbb{R}^{2n}$. The $n$ constraints corresponding to the upper bounds, $b_{kj}^>$, have weights $W^>_{kij} = 1$ where $i,j \in \{1 \ldots n \} \wedge i=j$, $0$ for $i \ne j$, and biases, $v^>_{kj} = - b_{kj}^>$. The corresponding entries for the lower bounds are $W^<_{kij} = -1$ where $i,j \in \{1 \ldots n \} \wedge i=j$, $0$ for $i \ne j$, and, $v^<_{kj} = b_{kj}^<$. We then have:
\begin{equation}
\mathbf{W}_k =
\begin{pmatrix}
\mathbf{W}_k^> \\
\mathbf{W}_k^<
\end{pmatrix}, \;
\mathbf{v}_k =
\begin{pmatrix}
\mathbf{v}_k^> \\
\mathbf{v}_k^<
\end{pmatrix}.
\end{equation}
The region of space composed of points, $\mathbf{x}$, that satisfy all of the inequalities corresponds to the interior of hypercube, $H_k$. If one or more of the inequalities is unsatisfied, the point lies on the exterior of $H_k$.

\subsection{Linear Matrix Inequalities Directly Specify RELU Layers in a Neural Network}

This system of inequalities should look suspiciously like the set of equations governing the input to a hidden layer within a neural network, $\mathbf{W}_k \mathbf{x} + \mathbf{v}_k$. And, in fact, using a ReLU activation function, the output of said layer would take the form:
\begin{equation}
\mathbf{z} = \text{ReLU} \left( \mathbf{W}_k \mathbf{x} + \mathbf{v}_k \right),
\end{equation}
where $\text{ReLU}(x) = x \, \theta(x)$ with $\theta(x)$ being the Heaviside step function. Note that by $\mathbf{z} = \text{ReLU}(.)$, we mean to imply the element-wise application of the ReLU function.

The nonlinearity of the ReLU function is similar to the boolean behavior of an inequality: yielding zero if the corresponding inequality is satisfied, and a finite value if it is not. Further, the sum, $\sum_i z_i$, will take on a value of zero if \emph{all} of the constraints are satisfied, that is, if a point lies within the interior of the corresponding hypercube, and a finite value if it does not. Let's, then, define a function:
\begin{equation}
\Theta_k(\mathbf{x}) = \sum_i \text{ReLU} \left( \mathbf{W}_k \mathbf{x} + \mathbf{v}_k \right) \big |_i,
\end{equation}
which indicates whether a point, $\mathbf{x}$, lies within the bounds of a hypercube, $H_k$. This function has a simple, one-to-one, correspondence with a neural network that has $n$ input nodes, represented by $\mathbf{x}$, and $2n$ hidden layer nodes, one corresponding to each constraint. The output of the hidden layer nodes is, then, summed, \emph{i.e.}, fed into a second hidden layer with a single node having a ReLU activation function; uniform weights, $\mathbf{1}$; and bias 0.

\begin{figure}
\center
\includegraphics[width=3.125in]{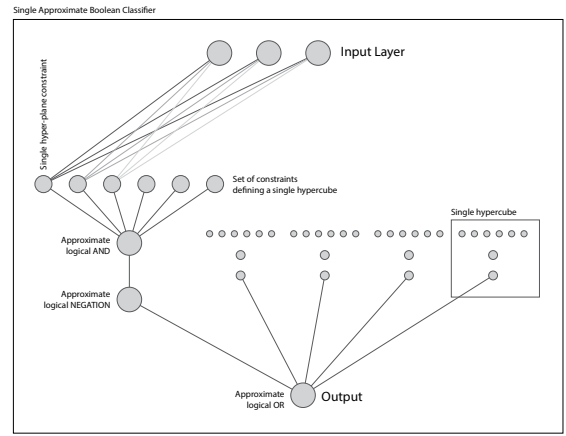} \\
\caption{A  schematic illustrating the realization of a neural network architecture generated by the constructive learning algorithm on the Iris training data set.  Nodes represent variables in input, hidden, or output layers while edges represent the multiplicative action of rows in the weight matrix of a given layer.  Bias terms are not represented in this diagram.  The realized neural network architecture spatially organizes nodes that represent individual hypercubes in the topological covering as collections of nodes.  Each node and layer can be directly linked to an aspect of the underlying topological cover of the data, resulting in an interpretable deep neural network.}
\label{fig:network-topology}
\end{figure}

If there are $K^c$ hypercubes that cover a region in parameter space corresponding to classification category, $c$, then there are $K^c$ corresponding functions, $\Theta^c_k$, each of which, in turn, corresponds to an independent hidden layer topology with uniquely defined weights. These $K^c$ outputs can then be combined in such a way as to yield $\Theta^c(\mathbf{x}) \sim 0$ if the point, $\mathbf{x}$, lies on the exterior of \emph{all} $K^c$ hypercubes, and $\Theta^c(\mathbf{x}) \sim 1$ if it lies within one of them:
\begin{equation}
\Theta^c(\mathbf{x}) = \sum_k 1 - \Theta_k^c(\mathbf{x}).
\end{equation}
This corresponds to an indication by the corresponding neural network of class membership. The $|C|$ outputs, $\Theta^c(\mathbf{x})$, where $|.|$ indicates cardinality, can then be combined with a softmax function to yield a traditional, one-hot neural network. The resulting neural network has the relatively shallow topology indicated in Fig. \ref{fig:network-topology}, where the nodes and connections corresponding to a classifier for a single category are illustrated. The outputs of each classifier are then combined, as suggested just above, using a softmax function.

Notice that there is some ambiguity in the resulting classification, indicated by the approximate equalities, $\Theta^c(\mathbf{x}) \sim 0$ and $\Theta^c(\mathbf{x}) \sim 1$, above. This ambiguity results from the fact that a logical negation is necessary; that is, the point $\mathbf{x}$ must be determined to lie within \emph{at least} one hypercube, which is to say that it must \emph{not} lie outside of all of the corresponding hypercubes. In the definition of the ReLU function, however, $\theta(x)$ yields a value of 0 both for values of $x<0$ (points on the interior of a hypercube) and for values of $x=0$ (points on, or infinitesimally close, to a hypercube boundary from the exterior). This ambiguity finds its way into $\Theta^c(\mathbf{x})$. What is needed, instead, is a function, $\Theta^c(\mathbf{x}; \mathbf{B}) = \Theta^c(|\mathbf{x} - \mathbf{B}|)$, with the following behavior:
\begin{equation}
\lim_{|\mathbf{x} - \mathbf{B}| \uparrow 0} \Theta^c(|\mathbf{x} - \mathbf{B}|) = 0 \text{ while } \lim_{|\mathbf{x} - \mathbf{B}| \downarrow 0} \Theta^c(|\mathbf{x} - \mathbf{B}|) = 1,
\end{equation}
where $|\mathbf{x} - \mathbf{B}|$ indicates a distance to the nearest boundary, and $\uparrow$ and $\downarrow$ indicate an approach to that boundary from the hypercube's interior and exterior, respectively. This can effectively be achieved with a minor change to our previous definition of $\Theta_k(\mathbf{x})$, by substituting:
\begin{equation}
\Theta^c(\mathbf{x}; \epsilon) = \sum_k 1 - \Theta_k^c(\mathbf{x}) / \epsilon,
\end{equation}
where $\epsilon$ represents a normalized length scale, assuming the data, itself, has been normalized. In the limit that $\epsilon \rightarrow 0$, the desired functional behavior is achieved. We note that for finite values of $\epsilon$, say $\epsilon = l$, the boundary of the hypercubes is essentially softened over a length scale, $l$. In some cases this has yielded better accuracy with respect to the evaluation data, effectively providing a smoother, more natural generalization of the training data.

\section{Numerical Results}

The performance of correct-by-construction neural networks, developed using the bisection algorithm described above, was compared to standard neural networks using a number of simple benchmark datasets. As noted in the introduction, a direct comparison is hindered by a couple of challenges associated with the standard approach to neural network development. These challenges highlight potential benefits of the approach that's been laid out herein. Both represent ambiguous areas in traditional network development and training. The first arises in the determination of an appropriate neural network topology. Traditionally, this is not learned as part of the training process, but must be specified statically beforehand. Yet, there is no \emph{a priori} rule for specifying network topology. It's historically been more art and experimentation than science. For the purposes of comparison here, many different topologies were explored ranging from 8 to 128 nodes in each of between one and three hidden layers. For the simple classification problems explored here, neither variable appeared to affect outcomes strongly. As such, all reported accuracies were extracted from a network with a single hidden layer of 32 nodes.

The second ambiguity concerns a determination of when a network has been trained ``enough''. In this particular study, evaluation accuracy appeared to depend far more strongly on the randomized weights initially assigned to the network than on the training time provided to it. The initial results presented here are a bit heuristic as the number of training iterations for each benchmark was selected by `eye' when training performance appeared to plateau. Future work will included a quantitative metric for training closure. As can be seen in the last two rows of the second table, however, increasing training time by an order of magnitude had minimal impact improving accuracy.

Notably, both of these ambiguities are addressed by the correct-by-construction algorithm. The determination of a hypercube covering uniquely specifies a resulting neural network topology. Additionally, the bisection algorithm simply runs to a deterministic completion, \emph{i.e.}, to the point that all training data are correctly classified, or a minimal scale has been reached, removing the ambiguity surrounding training time. Further, the performance of the bisection algorithm is far more repeatable than that of a traditionally trained neural network, especially when the number of data points available for training is relatively small. 

Quantitative testing was performed on a number of relatively simple, standard classification test problems. The first of these was the Iris Data Set \cite{Dua2017}. The Iris Data Set contains measurements of sepal length, sepal width, petal length, and petal width for 50 flower specimens from each of three iris species (for a total of 150 data points). A principle component analysis (PCA) was first done on these four parameters. Three different classification problems were then attempted using the first two, three, and four PCA directions. In each case a random selection of 70\% of the points was used for training and the remaining 30\% were used for evaluation. This process was replicated 20 times as the random selection of training and evaluation sets --- as well as stochastic processes within the training algorithms --- can impact performance. The results are reported in the tables below, first for the correct-by-construction algorithm, then for a traditionally-trained neural network.

There are two critical points to note when comparing the two. First, the bisection algorithm is able to achieve roughly 96\% accuracy when using two, three, or four PCA directions as compared to the roughly 75\% mean accuracy achieved by a traditionally trained neural network. Second, this accuracy is repeatably achieved by the correct-by-construction algorithm, with standard deviations of only 3\% or so from run to run. By comparison, the traditionally trained network has much more scatter shot performance, with a typical standard deviation of between 19\% and 24\% between runs.

A similar comparison was made using a Wine Data Set \cite{Forina}. In this data set, thirteen chemical and physical measurements were reported for each of 178 different wines, each wine an example of one of three different varietals. Measurements included such things as the alcohol content, non-flavanoid phenol levels, and visual hue. PCA was, again, done on the measurements before classification. The Hypercube Bisection Algorithm was able to achieve a mean accuracy of 87\% with a standard deviation of 4.4\%. The traditionally trained neural network yielded a mean accuracy of 82\% with a standard deviation of 17\%. Again, the bisection algorithm yielded better accuracy with significantly better reproducibility. Note that this isn't meant to imply that a traditionally-trained neural network can't perform at better than 82\% accuracy on this classification problem, only that there is distribution of performance outcomes, the mean of which is characterized by an 82\% accuracy. 

Finally, the two approaches were compared using the MNIST Digits data set. Each member of the data set represents a 28 x 28 pixel gray scale image of a handwritten digit, resulting in a total of 784 potential input values. In this comparison, rather than using a convolutional approach, PCA was performed on the data set and the first ten (10) PCA directions selected as inputs. Comparisons were made using both 200 and 2,000 point subsets of the data. Using 200 points, the mean accuracy of the bisection algorithm and traditionally trained network were fairly similar, 62\% and 64\%, respectively. Though, again, the bisection algorithm yielded more repeatable results, 5.0\% standard deviation versus 11\% standard deviation. With 2,000 training points the mean accuracies were again similar, in the mid- to upper-seventies, though the standard deviation of the bisection algorithm accuracy was only 1.9\%, while it was 8\%-10\% for the traditionally trained network.

These results suggest that for relatively simple classification problems the correct-by-construction, bisection-algorithm-based neural networks described above may require less computational complexity (for small numbers of training points), yield comparable or greater accuracies, and provide significantly more reproducible results than traditionally-trained neural networks. They also remove ambiguities surrounding network topology and training closure that are inherent in traditional network definition and training approaches.

\begin{table}[ht]
\begin{center}
\caption{Benchmarks of Hypercube Bisection Algorithm. Algorithm was tested using a random selection of 70\% of the data for training and 30\% for evaluation. Results represent the mean accuracy over 20 such tests.}
\begin{tabular}{lccc}
\hline
Dataset & Accuracy & Std Dev & Clock Time \\ \hline
Iris 2D PCA (150 pts) & 95\% & 3.6\% & 32ms \\
Iris 3D PCA (150 pts) & 96\% & 2.8\% & 38ms \\
Iris 4D PCA (150 pts) & 96\% & 2.3\% & 45ms \\ \hline
Wine PCA (178 pts) & 87\% & 4.4\% & 199ms \\ \hline
MNIST 10D PCA (200 pts) & 62\% & 5.0\% & 512ms \\
MNIST 10D PCA (2000 pts) & 76\% & 1.9\% & 22s \\ \hline
\end{tabular}
\end{center}
\end{table}

\begin{table}[ht]
\begin{center}
\caption{Benchmarks for traditionally trained neural network. Algorithm was tested using a random selection of 70\% of the data for training and 30\% for evaluation. Results represent the mean accuracy over 20 such tests.}
\begin{tabular}{lccc}
\hline
Dataset & Accuracy & Std Dev & Clock Time \\ \hline
Iris 2D PCA (150 pts) & 73\% & 24\% & 87ms \\
Iris 3D PCA (150 pts) & 77\% & 22\% & 85ms \\
Iris 4D PCA (150 pts) & 75\% & 19\% & 85ms \\ \hline
Wine PCA (178 pts) & 82\% & 17\% & 123ms \\ \hline
MNIST 10D PCA (200 pts) & 64\% & 11\% & 124ms \\
MNIST 10D PCA (2000 pts) & 75\% & 8\% & 330ms \\
MNIST 10D PCA (2000 pts) & 79\% & 10\% & 4.04s \\ \hline
\end{tabular}
\end{center}
\end{table}

\subsection{Conclusions}
In summary, deep learning algorithms rely on iterative and corrective refinement of a collection of model parameters and hyper-parameters to obtain a sufficiently accurate learning representation.  Such approaches require intensive training and extensive computational resources. 

In this paper we presented a novel constructive approach for training deep neural networks using geometric methods. The key insight is a fundamental relationship between linear matrix inequalities and their ability to bound the shape of data, and the rectified linear unit (ReLU) activation function employed in modern neural networks.  We show that the constructive algorithm is able to train a deep neural network classifier in one shot and show it achieves equal or superior levels of training and test classification accuracy with far less training time, in classical machine learning classification benchmark datasets.   In particular, we motivate the use of adaptive cover sizing approaches for generating data representations and their corresponding neural network representations.  In the context of topological cover learning, we revisit the age old trade-off between model complexity (as quantified by the number of elements in the data cover) and generalizability.  Our algorithm is the first in a new class of algorithms for constructive deep learning; future work will investigate Reeb graph and Morse theory methods for data shape decomposition and neural network parameterization. 



\end{document}